\documentclass[sigconf,authorversion,nonacm]{acmart}

\usepackage{xcolor}
\newcommand{\PH}[1]{{#1}}
\usepackage{multirow}
\usepackage{subcaption}
\usepackage{tabularx}
\usepackage{caption}
\usepackage{booktabs}
\usepackage{float}
\usepackage{placeins}
\usepackage{array}
\usepackage{enumitem} 
\usepackage{setspace} 
\usepackage{balance}  
\usepackage[ruled,linesnumbered,vlined]{algorithm2e}

\SetCommentSty{itshape}          
\SetKwComment{tcp}{$\triangleright$\ }{}   
\newcolumntype{C}{>{\centering\arraybackslash}X} 
\AtBeginDocument{%
  \providecommand\BibTeX{{%
    \normalfont B\kern-0.5em{\scshape i\kern-0.25em b}\kern-0.8em\TeX}}}

\setcopyright{acmcopyright}
\copyrightyear{2027}
\acmYear{2027}
\acmDOI{XXXXXXX.XXXXXXX}

\begin{document}

\title{JAPE: Joint Anomaly Prediction and Intrinsic Explanation in Multivariate Time Series}

\author{Yian Wei}
\affiliation{%
  \institution{Zhejiang University}
  \city{}
  \country{}
}
\email{yianwei@zju.edu.cn}

\author{Yuanyuan Yao}
\affiliation{%
  \institution{Zhejiang University}
  \city{}
  \country{}
}
\email{yoyoyao@zju.edu.cn}

\author{Lu Chen}
\affiliation{%
  \institution{Zhejiang University}
  \city{}
  \country{}
}
\email{luchen@zju.edu.cn}

\author{Xiangmin Zhou}
\affiliation{%
  \institution{School of Computing Technologies, RMIT University}
  \city{}
  \country{}
}
\email{xiangmin.zhou@rmit.edu.au}

\author{Tianyi Li}
\affiliation{%
  \institution{Aalborg University}
  \city{}
  \country{}
}
\email{tianyi@cs.aau.dk}

\begin{abstract}
Multivariate time-series anomaly prediction aims to identify whether and when anomalies will occur over a future horizon from historical observations. Existing methods primarily characterize anomalies as deviations in future numerical values, which may overlook subtle dependency changes induced by weak anomaly precursors and provide no native variable-level explanation together with the alert. To bridge these gaps, we propose JAPE, a \textbf{\underline{J}}oint
\textbf{\underline{A}}nomaly \textbf{\underline{P}}rediction and \textbf{\underline{E}}xplanation framework that lifts anomaly prediction from numerical-deviation modeling to dependency-structure modeling. \textbf{JAPE} is the first anomaly prediction framework to explicitly model evolving dependency structures for both point-wise alerting and native variable-level explanation. Specifically, \textbf{JAPE} (i) proposes a {Decoupled Spatio-Temporal Representation (DSTR)} backbone that decouples temporal and spatial modeling and captures lag-aware dependencies via learnable lag aggregation, thereby perceiving structural precursors before numerical deviations emerge; (ii) designs a dual-view alerting mechanism that fuses numerical forecasts with evolving dependency graphs for point-wise anomaly prediction, capturing structural evidence even under subtle numerical deviations; and (iii) presents {Native Predictive Explanation (NPE)}, which directly reuses the predicted dependency graphs to rank variables by structural deviations without additional models or training. Extensive experiments on five real-world benchmarks across three prediction horizons demonstrate that \textbf{JAPE} improves average F1 and AUC-PR by 19.7\% and 41.3\%, respectively, while improving explainability with 26.6\% gain in MRR.Å
\end{abstract}



\maketitle


\section{Introduction}

In modern cloud-native architectures and large-scale distributed systems, continuous monitoring of multivariate time series (MTS) is essential for maintaining system reliability and service availability. Traditional anomaly detection methods~\cite{su2019omnianomaly,audibert2020usad,xu2022anomalytransformer} identify abnormal behaviors only after faults have already occurred, providing reactive alerts that are often too late to prevent service degradation or operational losses. As system downtime becomes increasingly costly, there is a growing demand for proactive monitoring capabilities. This has led to a shift from post-hoc anomaly detection toward anomaly prediction, which aims to anticipate future anomalies before their occurrence by leveraging historical observations and temporal patterns: early studies, such as \textit{PoA}~\cite{jhin2023pad}, detect precursors indicating whether an anomaly is imminent, while later work further pinpoints abnormal time steps over an explicit future horizon~\cite{you2024anomaly}. However, anomaly prediction is more challenging than anomaly detection. Beyond identifying an upcoming anomaly, the anomaly prediction task must accurately predict its onset time and issue early point-wise alerts for proactive intervention. 

Existing approaches can be broadly divided into two paradigms. The first paradigm adopts an unsupervised formulation that combines a forecasting model~\cite{nie2023patchtst,liu2024itransformer} with reconstruction-based anomaly detection~\cite{xu2022anomalytransformer,audibert2020usad} to anticipate future anomalies~\cite{tfattn2025,fcm2025}. For example, \textit{TranAP}~\cite{tfattn2025} first learns forecasting and reconstruction models from entirely normal data; it then forecasts future series from historical observations, reconstructs the predicted series with the learned model, and flags time steps whose reconstruction error exceeds a predefined threshold as potential future anomalies. Going a step further along this line, \textit{FCM}~\cite{fcm2025} leverages the forecasted future context to amplify weak anomaly precursors within the observation window. While effective for pronounced numerical deviations, such methods are fundamentally optimized to model normal dynamics, so weak anomaly precursors are often overwhelmed by dominant normal patterns and remain unnoticed until the numerical deviation becomes pronounced. The second paradigm adopts a self-supervised formulation that injects synthetic pseudo-anomalies into normal data to construct auxiliary prediction objectives. For example, \textit{A2P}~\cite{park2025a2p} injects multi-form pseudo-anomalies through a learnable prompt pool, enabling the model to learn anomaly-related patterns during training. Recent work has also explored supervised anomaly prediction using a limited
number of real anomaly labels. \textit{F2A}~\cite{f2a2025}, for example,
adapts a pretrained model with a joint forecast--anomaly loss and fuses
retrieved relevant horizons through retrieval augmentation to improve
cross-system generalization. Yet whether the training signal comes from clean normal data, injected pseudo-anomalies, or a pretrained model, these methods are highly consistent in their modeling perspective: they all characterize anomalies essentially as deviations on the future numerical series, and raise alerts accordingly.

\begin{figure}[!t]
  \centering

    \begin{subfigure}[t]{0.49\columnwidth}
        \centering
        \includegraphics[width=\linewidth]{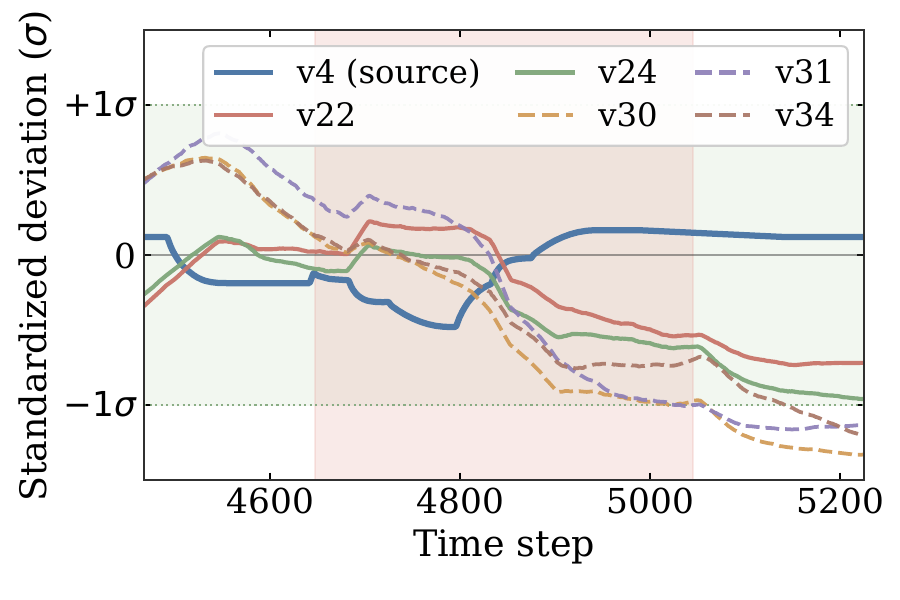}
        \caption{Per-variable trends}
        \label{fig:motiv_a}
    \end{subfigure}
    \hfill
    \begin{subfigure}[t]{0.49\columnwidth}
        \centering
        \includegraphics[width=\linewidth]{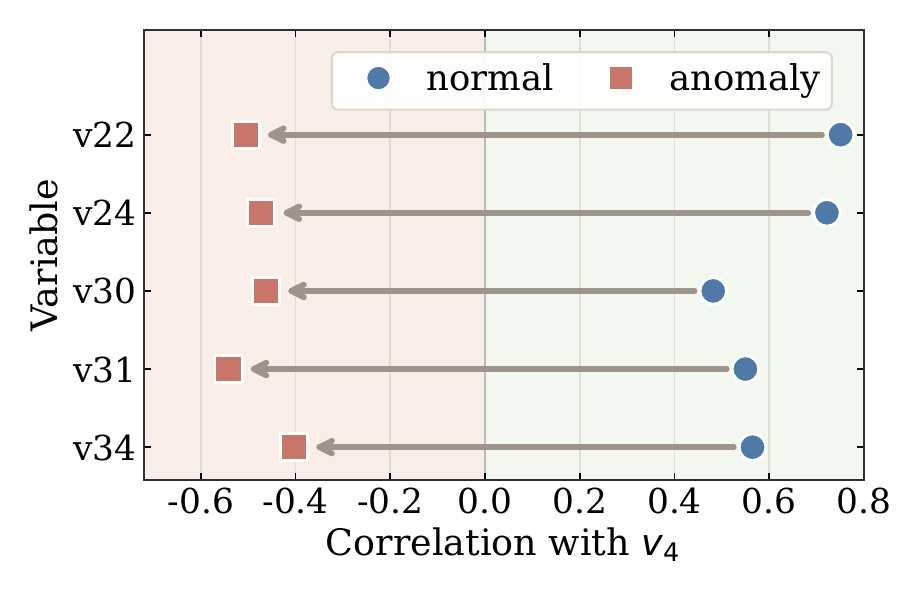}
        \caption{Correlation shifts }
        \label{fig:motiv_b}
    \end{subfigure}
  \caption{A case study on SMD (machine-1-6).}
  \vspace{-5mm}
  \label{fig:motivation}
  
\end{figure}

Existing value-centric approaches overlook the structural evidence carried by inter-variable relationships. In distributed systems, anomalies may manifest not only as changes in monitored signal values but also as structural changes in variable dependencies. Figure~\ref{fig:motivation} illustrates this phenomenon using an anomalous interval from the SMD dataset. Figure~\ref{fig:motiv_a} shows the temporal behaviors of an anomalous variable, \(v_4\), and several related variables around the anomaly interval. As shown in Figure~\ref{fig:motiv_b}, the observed change is not merely a fluctuation in correlation strength but a clear structural reorganization: variables that are predominantly positively correlated with \(v_4\) under normal conditions become mostly negatively correlated during the anomaly, resulting in a substantial shift in the overall dependency pattern. This observation suggests that inter-variable relationships provide meaningful structural evidence for anomaly-related system changes. Therefore, explicitly modeling evolving dependency structures is essential for both anomaly prediction and variable-level explanation. Consequently, existing value-centric approaches suffer from the following three fundamental limitations: {\textbf{\emph{L1: Delayed detection.}} Under numerical forecasting objectives, weak dependency changes are often absorbed as normal fluctuations and overwhelmed by dominant normal patterns, causing anomaly precursors to remain under-emphasized and alerts to lag behind the progression of faults.
\textbf{\emph{L2: Limited anomaly discriminability.}} During anomaly propagation, anomaly-source variable and other variables often exhibit similar numerical deviations. This challenge is exacerbated when source changes are subtle, making it difficult for value-centric methods to distinguish anomaly sources from propagated responses and reducing prediction accuracy.}
\textbf{\emph{L3: Lack of intrinsic explainability.}}
Without explicitly modeling dependency-structure evolution, existing methods detect anomalies but cannot directly identify the variables involved, requiring separate post-hoc analysis for interpretation.

To address these three limitations, we propose \textbf{JAPE}, a Joint Anomaly Prediction and Explanation framework that lifts anomaly prediction from numerical-deviation modeling to dependency-structure modeling, {trained under a practical setting where a small number of anomaly labels are available.}

\textbf{Decoupled Spatio-Temporal Representation (DSTR)}. To address \emph{\textbf{L1}}, we propose the DSTR backbone that explicitly separates temporal evolution from inter-variable dependency modeling. Instead of jointly learning temporal dynamics and structural dependencies in a single coupled operation, DSTR models them along two dedicated axes: a temporal axis for modeling the evolution of individual variables and a spatial axis for modeling inter-variable dependencies. Along the spatial axis, \textbf{JAPE} employs a \emph{lag-aware directional contrast} function together with a \emph{learnable lag aggregation} mechanism to capture evolving lead--lag dependency patterns. By modeling dependency evolution independently of temporal dynamics, DSTR preserves weak structural precursors before they manifest as observable numerical deviations.

\textbf{Dual-View Alerting Mechanism.}
To address \emph{\textbf{L2}}, \textbf{JAPE} complements the numerical view provided by
the predicted series $\widehat{\mathbf{X}}$ with the structural view provided
by the evolving dependency graph $\mathbf{A}$. The numerical view captures the
observable responses of individual variables, whereas the structural view
explicitly characterizes source-related changes in their pairwise relative influence.
The two views are adaptively fused through cross-attention, enabling \textbf{JAPE} to exploit
source-aware structural evidence even when the source variable exhibits only a
subtle numerical deviation, thereby improving anomaly discrimination.

\textbf{Native Predictive Explanation (NPE).}
To mitigate \emph{\textbf{L3}}, we reuse the dependency graph generated during
prediction to provide native explanations for anomaly alerts. When an anomaly
is detected, \textbf{JAPE} ranks variables according to their structural deviations from
the normal dependency pattern, thereby identifying those most associated with
the predicted anomaly. Because the dependency graph is already constructed
during inference, NPE requires neither additional training nor another model
forward pass, eliminating the need for a separate post-hoc diagnostic pipeline.

To the best of our knowledge, \textbf{JAPE} is the first anomaly prediction framework that explicitly models evolving dependency structures to jointly achieve {early point-wise alerting and native predictive explanation.} The contributions are summarized as follows. 

\begin{itemize}[topsep=0pt,itemsep=0pt,parsep=0pt,partopsep=0pt,leftmargin=*]
    \item We develop the Decoupled Spatio-Temporal Representation (DSTR) backbone, which disentangles temporal evolution from dependency modeling to preserve weak structural precursors via dynamic directed dependency graphs.
    \item We propose a Dual-View Alerting Mechanism that fuses
    numerical forecasts with dynamic dependency structures, enabling \textbf{JAPE} to
    exploit source-aware structural evidence when numerical responses remain subtle
    and thereby improving point-wise anomaly discrimination.
    \item We propose Native Predictive Explanation (NPE), which reuses the graph already produced during prediction and ranks variables via a Graph Deviation Score (GDS) against a normal baseline, yielding explanation at zero additional cost.
    \item We conduct extensive experiments on five datasets across three forecasting horizons to demonstrate the effectiveness of our proposed framework. Compared with strong baselines, \textbf{JAPE} achieves 19.7\% and 41.3\% average gains in F1 and AUC-PR under strict point-wise evaluation, alongside a 26.6\% MRR improvement for variable-level explanation.

\end{itemize}

\section{Preliminaries}
\label{sec:prelim}


{In this section, we first describe the multivariate time-series formulation and then formally define the anomaly prediction task.}

\begin{definition}[Multivariate Time Series]
A multivariate time series with $V$ variables is represented as a sequence of
observations over consecutive timestamps. Given a historical window of length
$L$, the time-series segment is defined as $\mathbf{X} =[\mathbf{x}_{t-L+1},\dots,\mathbf{x}_{t}] \in\mathbb{R}^{L\times V},$ where $\mathbf{x}_{t}\in\mathbb{R}^{V}$ denotes the observation vector at timestamp $t$, and each dimension corresponds to one variable.
\end{definition}

\begin{figure*}[!t]
  \centering
  \includegraphics[width=\textwidth]{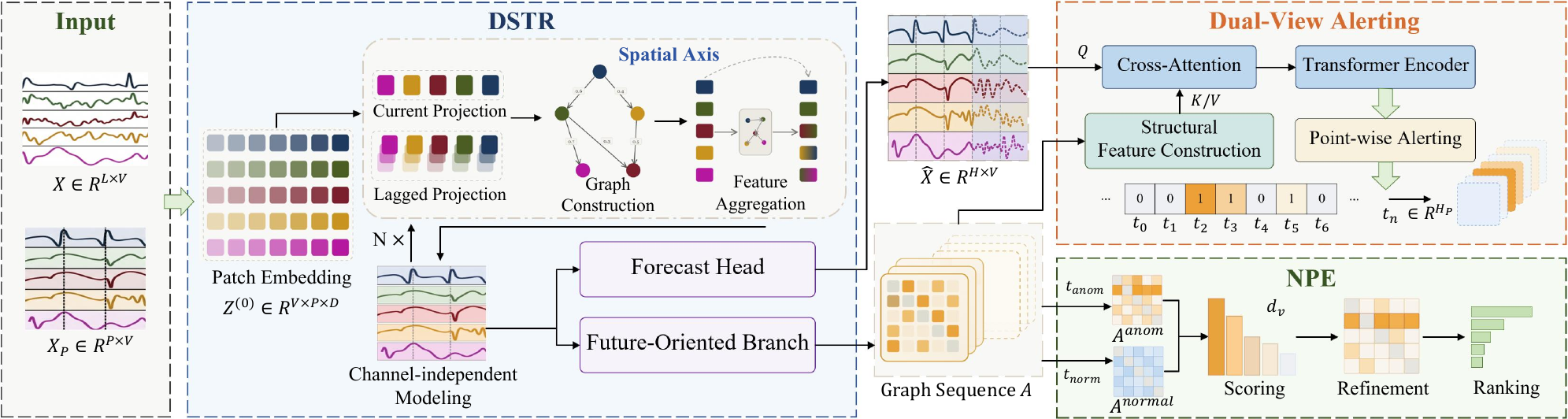}
  \vspace{-4mm}
  \caption{Overall architecture of JAPE.}
  \label{fig:arch}
   \vspace{-3mm}
\end{figure*}

\begin{definition}[Anomaly Prediction]
Given a historical observation window $\mathbf{X}$ of a multivariate time series and a prediction horizon $H$, anomaly prediction aims to estimate the future anomaly states over the
horizon. The corresponding anomaly labels are defined as $\mathbf{Y} = [y_{t+1},\dots,y_{t+H}] \in\{0,1\}^{H},$ where $y_{t+h}=1$ indicates that an anomaly occurs at timestamp $t+h$. The objective is to learn a mapping $f:\mathbf{X}\rightarrow\mathbf{P},$ where $\mathbf{P}\in[0,1]^H$ denotes the predicted probabilities of the corresponding anomaly labels in $\mathbf{Y}$
\end{definition}

In addition to future anomaly prediction, we further consider variable-level explanation of the predicted anomalies. Given an anomalous prediction, the goal is to identify variables that are most relevant to the anomaly occurrence. The explanation result is represented as a ranking over the $V$ variables.

\section{Method}
\label{sec:method}

In this section, we introduce the proposed \textbf{JAPE} framework. Figure~\ref{fig:arch} illustrates the overall architecture of \textbf{JAPE}, which consists of three key components: 1) \textbf{Decoupled Spatio-Temporal Representation (DSTR)}: It jointly models temporal dynamics and evolving inter-variable dependencies to produce both future numerical forecasts and dynamic directed dependency graphs; 2) \textbf{Dual-View Alerting Mechanism}: It integrates the numerical forecasting view and the structural dependency view to estimate point-wise anomaly probabilities; and 3) \textbf{Native Predictive Explanation (NPE)}: Once an anomaly is
detected, this component reuses the dependency graphs generated during
forecasting and computes Graph Deviation Scores (GDS) to identify the
variables most associated with the predicted anomaly. We next describe each component in detail.

\subsection{Decoupled Spatio-Temporal Representation (DSTR)}
\label{sec:dstr}
Anomalies in multivariate time series are often preceded by subtle changes in inter-variable dependencies rather than apparent numerical deviations.{A natural solution is to jointly model temporal evolution and cross-variable interactions with spatio-temporal attention~\cite{zhang2023crossformer}.} However, relations derived from current-token similarities mainly capture synchronous associations and fail to characterize lagged directional effects. Furthermore, jointly encoding temporal and variable interactions allows dominant temporal patterns to mask weak structural changes, which are easily overlooked under MSE optimization before numerical anomalies arise. To address this issue, DSTR decouples spatio-temporal modeling into
two separate axes. After patch embedding, the spatial axis explicitly learns dynamic directed dependencies among variables through lag-aware asymmetric graph construction, while the temporal axis models the evolution of each variable independently. These two axes are applied alternately within each encoder layer. Furthermore, a future-oriented branch extends spatial dependency modeling into the prediction horizon, enabling DSTR to preserve evolving dependency structures for future anomaly prediction.

\subsubsection{Patch Embedding}

Given an input window
$\mathbf{X}\in\mathbb{R}^{L\times V}$, where $L$ is the historical
length and $V$ is the number of variables, each univariate series
$\mathbf{X}_{:,v}\in\mathbb{R}^{L}$ is divided into overlapping
patches. Let $L_p$ denote the patch length and $S_p$ the stride. The
patching process produces a patch sequence
$\mathbf{X}^{p}_{v}\in\mathbb{R}^{P\times L_p}$ for variable $v$,
where
$P=\left\lfloor(L-L_p)/S_p\right\rfloor+1$
is the number of patches. Each patch is mapped to a
$D$-dimensional token through a shared linear projection followed by
positional encoding. Thus, each variable is represented by $P$ patch tokens of dimension
$D$, forming
$\mathbf{Z}^{(0)}\in\mathbb{R}^{V\times P\times D}$, which is passed
through $N$ dual-axis encoder layers.

\subsubsection{Spatial Axis: Lag-Aware Directed Graph Learning}

The spatial axis aims to characterize directed influences among
variables and to preserve them explicitly as a graph, independently of
numerical magnitude. At patch position $i$ in encoder layer $\ell$, let
$\mathbf{Z}_{i}^{(\ell-1)}
=
\mathbf{Z}_{:,i,:}^{(\ell-1)}
\in\mathbb{R}^{V\times D}$ denote the input tokens of all variables.
Three learnable linear projections produce
$\mathbf{Q}_{i}^{(\ell)}$,
$\mathbf{K}_{i}^{(\ell)}$, and
$\mathbf{U}_{i}^{(\ell)}
\in\mathbb{R}^{V\times D}$, which are used to measure pairwise
influences across variables.

\noindent \textbf{Current and lagged projections.}
A score computed solely from the current patch captures only instantaneous association and fails to represent delayed influences. To model such lagged effects, we aggregate the preceding $K_{\max}$ patch representations via exponentially decayed weighting to construct a lagged representation for the current patch. Specifically, the lag weight is defined as
\begin{equation}
a_k = \frac{\exp[-\delta(k-1)]}{\sum_{r=1}^{K_{\max}} \exp[-\delta(r-1)]}, \quad \delta=\exp(\theta)>0,
\label{eq:lag-weight}
\end{equation}
where $\theta$ is a learnable parameter and a larger $\delta$ assigns higher weight to more recent patches. For any projection $\mathbf{M}_{i}^{(\ell)} \in \{\mathbf{Q}_{i}^{(\ell)}, \mathbf{K}_{i}^{(\ell)}, \\ \mathbf{U}_{i}^{(\ell)}\}$, its lagged counterpart is defined as the weighted aggregation of the corresponding projections from preceding patches:
\begin{equation}
\widetilde{\mathbf{M}}_{i}^{(\ell)} = \sum_{k=1}^{K_{\max}} a_k \mathbf{M}_{\max(1,i-k)}^{(\ell)},
\label{eq:lag-aggregation}
\end{equation}
where $\max(1,i-k)$ handles sequence boundaries by truncating indices below $1$ to the first valid patch. Consequently, $\widetilde{\mathbf{M}}_{i}^{(\ell)}$ denotes the set $\{\widetilde{\mathbf{Q}}_{i}^{(\ell)}, \widetilde{\mathbf{K}}_{i}^{(\ell)}, \widetilde{\mathbf{U}}_{i}^{(\ell)}\}$, corresponding to the lagged query, key, and value projections, respectively.

\noindent \textbf{Directed graph construction.}
Similarity-based attention~\cite{vaswani2017attention} does not
explicitly distinguish the two directions of an ordered variable pair.
We therefore introduce a lag-aware directional contrast score at each
patch position $i$. The directional score matrix
$\mathbf{S}_{i}^{(\ell)}\in\mathbb{R}^{V\times V}$ is defined as:
\begin{equation}
\mathbf{S}_{i}^{(\ell)}
=
\mathbf{Q}_{i}^{(\ell)}
\left(\widetilde{\mathbf{K}}_{i}^{(\ell)}\right)^{\top}
-
\mathbf{K}_{i}^{(\ell)}
\left(\widetilde{\mathbf{Q}}_{i}^{(\ell)}\right)^{\top}.
\label{eq:antisym}
\end{equation}
For an ordered pair $(u,v)$, $S_{i,uv}^{(\ell)}$ measures the difference
between the alignment of $u$'s current state with $v$'s lagged state
and that of $v$'s current state with $u$'s lagged state. A positive
value indicates directional evidence from $v$ to $u$: the past of
$v$ helps predict the present of $u$. For each target variable, we
retain the $K_g$ source variables with the strongest positive
dependency scores:
\begin{equation}
\mathbf{A}^{\mathrm{raw},(\ell)}_{i}
=
\operatorname{TopK}_{K_g}
\left(
\operatorname{ReLU}
\left(
\tanh(\mathbf{S}_{i}^{(\ell)})
\right)
\right),
\label{eq:graph-sparsification}
\end{equation}
where $\operatorname{TopK}_{K_g}$ retains the selected dependency
scores and sets the others to zero. The resulting
$\mathbf{A}^{\mathrm{raw},(\ell)}_{i}
\in\mathbb{R}^{V\times V}$
is normalized across the selected source variables for each target
variable, yielding
$\mathbf{A}_{i}^{(\ell)}\in\mathbb{R}^{V\times V}$.


\noindent\textbf{Graph-guided feature aggregation and graph output.}
The learned graph guides feature propagation and serves as the
structural representation for anomaly prediction. Specifically, each
variable aggregates historical features from its selected predecessors:
\begin{equation}
\mathbf{Z}_{i,\mathrm{spat}}^{(\ell)}
=
\mathbf{Z}_{i}^{(\ell-1)}
+
\lambda
\mathbf{A}_{i}^{(\ell)}
\widetilde{\mathbf{U}}_{i}^{(\ell)},
\end{equation}
where $\lambda$ is a learnable residual coefficient. This operation
injects dependency-aware information while preserving the original
representation through residual learning.

The spatial axis generates a graph at each patch position and encoder
layer. For each patch $i$, we average the normalized graphs across
layers as
$\mathbf{A}_{i}=\frac{1}{N}\sum_{\ell=1}^{N}
\mathbf{A}_{i}^{(\ell)}$.
The resulting sequence
$\{\mathbf{A}_{i}\}_{i=1}^{P}$ captures the evolution of
inter-variable dependencies over the historical window. The raw graphs
$\mathbf{A}_{i}^{\mathrm{raw},(\ell)}$ are averaged in the same manner
to preserve edge magnitudes.

\begin{table*}[t]
\centering
\caption{Structural Features Used by the Dual-View Alerting Mechanism}
\label{tab:structural-features}
\vspace{-3mm}

\begin{tabularx}{\textwidth}{
@{}
c|
l|
l|
l|
>{\raggedright\arraybackslash}l|
>{\raggedright\arraybackslash}X
@{}
}
\toprule
No. & Feature & Level & Graph Input & Statistic & Role \\
\midrule

1
& Mean Entropy
& Slice-level
& Normalized
& Avg.\ row entropy
& Dependency dispersion \\

2--3
& Top-$k$ Share
& Slice-level
& Normalized
& Avg.\ Top-$k$ sum, $k=1,3$
& Dependency concentration \\

4
& Outgoing Dominance
& Slice-level
& Normalized
& $\max_v s_v^{\mathrm{out}}/\sum_v s_v^{\mathrm{out}}$
& Variable dominance \\

5
& In--out Imbalance
& Slice-level
& Normalized
& Avg.\ $\lvert s_v^{\mathrm{in}}-s_v^{\mathrm{out}}\rvert$
& Directional imbalance \\

6--8
& Incoming Change
& Segment-level
& Unnormalized
& Max / avg.\ / std.\ $\Delta s_v^{\mathrm{in}}$
& Incoming redistribution \\

9--11
& Outgoing Change
& Segment-level
& Unnormalized
& Max / avg.\ / std.\ $\Delta s_v^{\mathrm{out}}$
& Outgoing redistribution \\

12--13
& Graph Energy
& Segment-level
& Unnormalized
& $E_{\mathrm{tail}},\ \Delta E$
& Energy level and shift \\

\bottomrule
\end{tabularx}
\end{table*}

\subsubsection{Temporal Axis: Channel-Independent Temporal Modeling}

Following the channel-independent design of
PatchTST~\cite{nie2023patchtst}, the temporal axis models the evolution
of each variable separately while sharing its parameters across
variables. For each variable $v$ at layer $\ell$, the temporal axis applies self-attention along the
patch sequence
$\mathbf{Z}_{v,:, \mathrm{spat}}^{(\ell)}
\in\mathbb{R}^{P\times D}$, without interaction
between different variables on this axis. This yields the layer output
$\mathbf{Z}^{(\ell)}\in\mathbb{R}^{V\times P\times D}$. This design keeps temporal evolution separate from the inter-variable
dependencies already captured by the preceding spatial axis, preserving
the explicit graph structure for subsequent alerting and explanation.We denote the final historical representation by
$\mathbf{Z}=\mathbf{Z}^{(N)}$.

\subsubsection{Future-Oriented Branch and Graph Concatenation}

While the historical spatial axis captures dependencies within the
observed window, anomaly detection also benefits from structural evidence
aligned with the future horizon, enabling a before-and-after perspective
at the time of alerting.

Let $H_p=\lfloor(H-L_p)/S_p\rfloor+1$ denote the number of future
patches under the same patching rule. The future-oriented branch maps a
stop-gradient copy of the final historical representation
$\mathbf{Z}^{(N)}$ to
$\widehat{\mathbf{Z}}_{\mathrm{fut}}
\in\mathbb{R}^{V\times H_p\times D}$. An independent spatial block then constructs
$\{\widehat{\mathbf{A}}_j\}_{j=1}^{H_p}$ at each future patch position
using the same lag-aware directional mechanism. The graph-enhanced
representation is decoded into an auxiliary forecast
$\widehat{\mathbf{X}}_{\mathrm{fut}}\in\mathbb{R}^{H\times V}$.

Both forecasting outputs are trained with the same MSE formulation.
For the main forecast, we use
$\mathcal{L}_{\mathrm{MSE}}
=\frac{1}{HV}\|\widehat{\mathbf{X}}-\mathbf{X}^{+}\|_{F}^{2}$.
For the future-oriented branch, the auxiliary objective is
$\mathcal{L}_{\mathrm{MSE}}^{\mathrm{fut}}
=\frac{1}{HV}\|\widehat{\mathbf{X}}_{\mathrm{fut}}
-\mathbf{X}^{+}\|_{F}^{2}$, which provides ground-truth supervision
for learning the future dependency graphs.

Finally, the historical and future graph sequences are concatenated
along the patch dimension to form
$\mathbf{A}\in\mathbb{R}^{(P+H_p)\times V\times V}$, which is passed
to the alerting module.

\subsection{Dual-View Alerting Mechanism}
\label{sec:head}

The Dual-View Alerting Mechanism estimates point-wise anomaly probabilities over the prediction horizon by jointly modeling future numerical evolution and evolving dependency structures. The numerical view captures the future trajectories of individual variables, whereas the structural view encodes the evolution of inter-variable dependencies produced by DSTR. Their combination enables anomaly prediction even when numerical precursors remain weak but structural patterns have already changed.

\noindent\textbf{Structural feature construction.} The dynamic dependency graphs generated by DSTR provide rich structural evidence for anomaly prediction. However, directly using adjacency matrices introduces variable-dependent input dimensions and redundant edge-level information. We therefore summarize each graph slice into a compact structural descriptor with a fixed dimensionality.

Let $T=P+H_p$ denote the total number of historical and future graph
slices. Let $s_v^{\mathrm{in}}$ and $s_v^{\mathrm{out}}$ denote the incoming
and outgoing strengths of variable $v$, respectively, and define the
graph energy at slice $t$ as
$E_t=\sum_{u,v}A^{\mathrm{raw}}_{t,uv}$.
Within each historical or future graph segment, the final $\tau$
portion is treated as the tail, and $\Delta$ denotes the change from
the preceding portion to the tail. Accordingly,
$E_{\mathrm{tail}}$ is the mean graph energy over the tail. Each graph slice is represented by a 13-dimensional structural
descriptor consisting of two complementary components: five
slice-level statistics computed from the normalized graph and eight
segment-level statistics computed from the corresponding unnormalized
graph sequence, as summarized in
Table~\ref{tab:structural-features}. The segment-level statistics are
appended to every graph slice within the corresponding historical or
future segment. All structural descriptors are organized
chronologically as
$\mathbf{F}^{\mathrm{str}}\in\mathbb{R}^{T\times13}$,
preserving both the dependency pattern of each graph slice and the
structural evolution of its surrounding segment.

\noindent\textbf{Cross-attention fusion.} We linearly project the predicted vectors at the $H$ future steps into
numerical tokens $\mathbf{H}^{x}\in\mathbb{R}^{H\times D}$ and project
the structural descriptor matrix $\mathbf{F}^{\mathrm{str}}$ into
structural tokens $\mathbf{H}^{g}\in\mathbb{R}^{T\times D}$.
Positional information is added to both sequences, and the structural
tokens additionally encode whether they originate from the historical
or future graph segment. {Using the numerical tokens as queries and the structural tokens as
keys and values, cross-view fusion is computed as}
\begin{equation}
\mathbf{C}
=
\operatorname{Softmax}\!\left(
\frac{
(\mathbf{H}^{x}\mathbf{W}_{Q})
(\mathbf{H}^{g}\mathbf{W}_{K})^{\top}
}{
\sqrt{D}
}
\right)
\mathbf{H}^{g}\mathbf{W}_{V},
\label{eq:dual-view-fusion}
\end{equation}
where $\mathbf{W}_{Q}$, $\mathbf{W}_{K}$, and $\mathbf{W}_{V}$ are
learnable projections. The resulting
$\mathbf{C}\in\mathbb{R}^{H\times D}$ is fused with the numerical
tokens through a residual connection and layer normalization.

\noindent\textbf{Point-wise alerting}
The fused representations are processed by a temporal
Transformer encoder to model interactions across the prediction
horizon. A linear output layer maps each encoded token to an anomaly
logit, yielding point-wise anomaly probabilities
$\mathbf{p}\in[0,1]^H$. During this stage, the DSTR backbone is frozen and only the alerting head is optimized using focal loss~\cite{lin2017focal}. 
The complete two-stage training procedure is described in Appendix~\ref{app:training}.

\subsection{Native Predictive Explanation}
\label{sec:npe}

The dependency graphs generated during anomaly prediction provide structural
evidence for identifying variables associated with future anomalies. Instead of
introducing an additional attribution model, NPE derives variable-level
explanations directly from the predicted dependency graphs by measuring their
deviations from normal dependency patterns. Specifically, we design a Graph
Deviation Score (GDS) that captures both immediate dependency changes and their
propagation along multi-hop dependencies.

For each alerted anomalous segment, we aggregate the predicted graph slices
within the segment into a segment-level dependency graph
$\mathbf{A}^{\mathrm{anom}}\in\mathbb{R}^{V\times V}$, where
$A_{uv}$ denotes the dependency from variable $v$ to variable $u$. A normal
reference graph $\mathbf{A}^{\mathrm{normal}}$ is constructed by aggregating
dependency graphs from normal reference windows. These normal windows are also
used to estimate variable-wise statistics for standardizing the deviation scores, reducing the
impact of heterogeneous dependency scales across variables.

The GDS first measures direct dependency changes between anomalous and normal
graphs. Let $[x]_{+}=\max(x,0)$. The direct deviation score of variable $v$ is
defined as

\begin{equation}
d^{\mathrm{dir}}_v
=
\sum_u
\left[
A^{\mathrm{anom}}_{uv}
-
A^{\mathrm{normal}}_{uv}
\right]_{+}
-
\beta
\sum_u
\left[
A^{\mathrm{anom}}_{vu}
-
A^{\mathrm{normal}}_{vu}
\right]_{+}.
\label{eq:direct-score}
\end{equation}

The first term measures strengthened outgoing dependencies, while the second
term penalizes strengthened incoming dependencies. This asymmetric formulation
highlights variables exhibiting upstream-related dependency changes during
anomalous periods. 

Direct dependency deviations capture local structural changes but may overlook
variables whose influence propagates through longer dependency chains. To
capture such structural propagation, we compare multi-hop dependency patterns
between anomalous and normal graphs. Specifically, Specifically, after row-normalizing each target variable's incoming dependency weights over its selected sources to obtain $\bar{\mathbf{A}}^{\mathrm{anom}}$ and $\bar{\mathbf{A}}^{\mathrm{normal}}$, the positive deviation of $k$-hop dependencies is computed as

\begin{equation}
\boldsymbol{\Delta}^{(k)}
=
\mathrm{ReLU}
\left(
(\bar{\mathbf{A}}^{\mathrm{anom}})^k
-
(\bar{\mathbf{A}}^{\mathrm{normal}})^k
\right),
\label{eq:multi-hop-deviation}
\end{equation}
where $\boldsymbol{\Delta}^{(k)}$ represents the changed dependency patterns
along paths of length $k$. The multi-hop deviation score is then defined as

\begin{equation}
d^{\mathrm{path}}_v
=
\sum_{k=2}^{K_{\mathrm{path}}}
w_k
\left(
\sum_u\Delta^{(k)}_{uv}
-
\beta
\sum_u\Delta^{(k)}_{vu}
\right),
\label{eq:path-score}
\end{equation}
where $w_k$ is a decay weight that reduces the contribution of longer
dependency paths. The direct and multi-hop scores are standardized via variable-wise statistics from normal reference windows, producing $\widetilde{\mathbf{d}}^{\mathrm{dir}}$ and $\widetilde{\mathbf{d}}^{\mathrm{path}}$. These complementary components are then combined to obtain the initial GDS representation:

\begin{equation}
\mathbf{g}^{(0)}
=
\widetilde{\mathbf{d}}^{\mathrm{dir}}
+
\omega
\widetilde{\mathbf{d}}^{\mathrm{path}},
\label{eq:gds-init}
\end{equation}
where $\omega$ controls the contribution of multi-hop dependency evidence. To
further incorporate normal dependency structures, we refine the scores through
residual propagation:

\begin{equation}
\mathbf{g}^{(r+1)}
=
\alpha_{\mathrm{gds}}\mathbf{g}^{(r)}
+
(1-\alpha_{\mathrm{gds}})
\frac{\mathbf{g}^{(r)}}
{\|\mathbf{g}^{(r)}\|_{\infty}+\epsilon}
\bar{\mathbf{A}}^{\mathrm{normal}},
\label{eq:gds-refinement}
\end{equation}
where $\alpha_{\mathrm{gds}}$ controls the balance between preserving the
original deviation evidence and incorporating dependency-aware refinement.

After $R$ refinement steps, the final explanation score is obtained as
$\mathbf{s}=\mathbf{g}^{(R)}$. Ranking variables according to $\mathbf{s}$
produces the variable-level explanation. Since NPE directly reuses the
dependency graphs generated during anomaly prediction, it requires no
additional attribution model, extra inference process, or root-cause
supervision.

\section{Experiments}
\label{sec:exp}
\begin{table*}[!t]
\centering
\scriptsize
\caption{Anomaly prediction performance (F1 / AUC-PR, \%) under
different prediction lengths $L_{out}$. Best results are in
\textbf{bold}, and second-best results are \underline{underlined}.}
\label{tab:main}
\vspace{-3mm}
\begingroup
\renewcommand{\arraystretch}{0.92}
\setlength{\aboverulesep}{0.30ex}
\setlength{\belowrulesep}{0.30ex}
\resizebox{\textwidth}{!}{
\begin{tabular}{c cc cc cc cc cc cc cc}
\toprule
\multirow{3}{*}{$L_{out}$}
& \multicolumn{2}{c}{\multirow{2}{*}{Model}}
& \multicolumn{10}{c}{Dataset (F1 / AUC-PR)}
& \multicolumn{2}{c}{Avg.} \\

\cmidrule(lr){4-13}
\cmidrule(lr){14-15}

& \multicolumn{2}{c}{}
& \multicolumn{2}{c}{SMD}
& \multicolumn{2}{c}{WADI}
& \multicolumn{2}{c}{MSL}
& \multicolumn{2}{c}{PSM}
& \multicolumn{2}{c}{EXATHLON}
& \multirow{2}{*}{F1}
& \multirow{2}{*}{PR} \\

\cmidrule(lr){2-3}
\cmidrule(lr){4-5}
\cmidrule(lr){6-7}
\cmidrule(lr){8-9}
\cmidrule(lr){10-11}
\cmidrule(lr){12-13}

& Forecast
& Detector
& F1 & PR
& F1 & PR
& F1 & PR
& F1 & PR
& F1 & PR
& & \\

\midrule
\multirow{8}{*}{50}
& \multicolumn{2}{c}{RED-F}
& \PH{13.8} & \PH{33.5} & \PH{11.9} & \PH{5.9} & \PH{19.1} & \PH{18.5} & \PH{43.5} & \PH{41.3} & \PH{40.2} & \PH{47.6} & \PH{25.7} & \PH{29.4} \\
& \multicolumn{2}{c}{FCM}
& \PH{16.5} & \PH{28.1} & \PH{11.3} & \PH{5.4} & \PH{23.9} & \PH{10.8} & \PH{45.4} & \PH{28.1} & \PH{52.3} & \PH{14.8} & \PH{29.9} & \PH{17.4} \\
\cmidrule(lr){2-15}
& \multirow{2}{*}{PatchTST}
& AT  & \PH{42.7} & \PH{43.0} & \PH{35.4} & \PH{33.4} & \PH{24.8} & \PH{16.2} & \PH{68.8} & \PH{81.8} & \PH{68.6} & \PH{75.2} & \PH{48.1} & \PH{49.9} \\
& & CAD & \PH{42.9} & \PH{42.6} & \PH{41.0} & \PH{28.1} & \PH{42.2} & \underline{\PH{30.9}} & \underline{\PH{70.6}} & \PH{79.6} & \underline{\PH{72.9}} & \PH{75.3} & \underline{\PH{53.9}} & \PH{51.3} \\
\cmidrule(lr){2-15}
& \multirow{2}{*}{iTrans.}
& AT  & \PH{41.6} & \PH{44.7} & \underline{\PH{51.4}} & \underline{\PH{49.3}} & \PH{23.4} & \PH{10.7} & \PH{69.6} & \PH{79.2} & \PH{71.0} & \PH{71.9} & \PH{51.4} & \PH{51.2} \\
& & CAD & \PH{44.6} & \PH{44.6} & \PH{40.3} & \PH{26.8} & \underline{\PH{46.9}} & \underline{\PH{30.9}} & \PH{66.1} & \PH{75.9} & \PH{71.4} & \PH{73.8} & \underline{\PH{53.9}} & \PH{50.0} \\
\cmidrule(lr){2-15}
& \multicolumn{2}{c}{A2P-Sup}
& \underline{\PH{46.2}} & \underline{\PH{44.9}} & \textbf{\PH{51.5}} & \PH{44.9} & \PH{24.0} & \PH{13.6} & \PH{68.8} & \underline{\PH{82.5}} & \PH{71.1} & \underline{\PH{76.9}} & \PH{52.3} & \underline{\PH{52.6}} \\
\cmidrule(lr){2-15}
& \multicolumn{2}{c}{\textbf{JAPE (Ours)}}
& \textbf{\PH{53.8}} & \textbf{\PH{54.6}} & \PH{42.5} & \textbf{\PH{62.8}} & \textbf{\PH{62.0}} & \textbf{\PH{68.1}} & \textbf{\PH{75.6}} & \textbf{\PH{91.1}} & \textbf{\PH{85.4}} & \textbf{\PH{91.6}} & \textbf{\PH{63.9}} & \textbf{\PH{73.6}} \\
\midrule
\multirow{8}{*}{100}
& \multicolumn{2}{c}{RED-F}
& \PH{14.1} & \PH{9.0} & \PH{11.4} & \PH{6.1} & \PH{19.1} & \PH{17.4} & \PH{43.4} & \PH{40.3} & \PH{41.2} & \PH{44.3} & \PH{25.8} & \PH{23.4} \\
& \multicolumn{2}{c}{FCM}
& \PH{14.4} & \PH{4.6} & \PH{11.6} & \PH{5.2} & \PH{23.9} & \PH{11.6} & \PH{45.3} & \PH{28.9} & \PH{50.1} & \PH{15.3} & \PH{29.1} & \PH{13.1} \\
\cmidrule(lr){2-15}
& \multirow{2}{*}{PatchTST}
& AT  & \underline{\PH{40.3}} & \PH{39.7} & \PH{48.8} & \PH{41.4} & \PH{26.1} & \PH{13.6} & \underline{\PH{69.0}} & \PH{79.0} & \PH{67.2} & \underline{\PH{72.6}} & \PH{50.3} & \PH{49.3} \\
& & CAD & \PH{37.7} & \PH{38.9} & \PH{47.8} & \PH{38.6} & \PH{44.2} & \underline{\PH{31.8}} & \PH{65.5} & \PH{75.6} & \underline{\PH{69.3}} & \PH{71.6} & \underline{\PH{52.9}} & \underline{\PH{51.3}} \\
\cmidrule(lr){2-15}
& \multirow{2}{*}{iTrans.}
& AT  & \PH{39.9} & \underline{\PH{40.2}} & \PH{49.2} & \PH{36.0} & \PH{25.4} & \PH{9.9} & \PH{65.5} & \PH{77.4} & \PH{65.8} & \PH{67.4} & \PH{49.2} & \PH{46.2} \\
& & CAD & \PH{36.8} & \PH{38.0} & \PH{46.6} & \PH{32.5} & \underline{\PH{45.5}} & \PH{30.1} & \PH{64.7} & \PH{76.0} & \PH{69.0} & \PH{70.9} & \PH{52.5} & \PH{49.5} \\
\cmidrule(lr){2-15}
& \multicolumn{2}{c}{A2P-Sup}
& \PH{37.1} & \PH{37.1} & \textbf{\PH{52.9}} & \underline{\PH{44.2}} & \PH{24.2} & \PH{14.3} & \PH{68.8} & \underline{\PH{80.6}} & \PH{61.2} & \PH{70.1} & \PH{48.8} & \PH{49.3} \\
\cmidrule(lr){2-15}
& \multicolumn{2}{c}{\textbf{JAPE (Ours)}}
& \textbf{\PH{52.5}} & \textbf{\PH{53.8}} & \underline{\PH{49.4}} & \textbf{\PH{62.0}} & \textbf{\PH{61.2}} & \textbf{\PH{66.1}} & \textbf{\PH{79.0}} & \textbf{\PH{88.3}} & \textbf{\PH{82.7}} & \textbf{\PH{88.1}} & \textbf{\PH{65.0}} & \textbf{\PH{71.7}} \\
\midrule
\multirow{8}{*}{200}
& \multicolumn{2}{c}{RED-F}
& \PH{13.7} & \PH{8.3} & \PH{11.4} & \PH{5.7} & \PH{18.9} & \PH{15.5} & \PH{43.3} & \PH{31.2} & \PH{40.1} & \PH{39.2} & \PH{25.5} & \PH{20.0} \\
& \multicolumn{2}{c}{FCM}
& \PH{13.6} & \PH{5.6} & \PH{11.5} & \PH{5.3} & \PH{23.9} & \PH{13.9} & \PH{44.9} & \PH{30.7} & \PH{46.6} & \PH{16.8} & \PH{28.1} & \PH{14.5} \\
\cmidrule(lr){2-15}
& \multirow{2}{*}{PatchTST}
& AT  & \PH{33.5} & \PH{34.1} & \underline{\PH{53.1}} & \underline{\PH{51.8}} & \PH{22.9} & \PH{18.5} & \textbf{\PH{67.8}} & \PH{77.4} & \PH{60.4} & \underline{\PH{67.3}} & \PH{47.5} & \underline{\PH{49.8}} \\
& & CAD & \PH{31.4} & \PH{32.7} & \PH{41.1} & \PH{33.2} & \PH{37.6} & \underline{\PH{27.4}} & \PH{64.9} & \PH{73.8} & \underline{\PH{63.4}} & \PH{64.7} & \PH{47.7} & \PH{46.4} \\
\cmidrule(lr){2-15}
& \multirow{2}{*}{iTrans.}
& AT  & \PH{33.1} & \PH{31.5} & \PH{44.9} & \PH{43.9} & \PH{10.8} & \PH{13.4} & \PH{67.0} & \PH{75.5} & \PH{56.0} & \PH{63.8} & \PH{42.4} & \PH{45.6} \\
& & CAD & \PH{34.8} & \PH{33.6} & \PH{49.6} & \PH{41.2} & \underline{\PH{43.5}} & \PH{27.0} & \PH{65.7} & \PH{74.9} & \PH{62.4} & \PH{64.1} & \underline{\PH{51.2}} & \PH{48.2} \\
\cmidrule(lr){2-15}
& \multicolumn{2}{c}{A2P-Sup}
& \underline{\PH{37.6}} & \underline{\PH{34.8}} & \underline{\PH{53.1}} & \PH{46.2} & \PH{28.6} & \PH{17.1} & \underline{\PH{67.7}} & \underline{\PH{79.2}} & \PH{55.2} & \PH{66.0} & \PH{48.4} & \PH{48.7} \\
\cmidrule(lr){2-15}
& \multicolumn{2}{c}{\textbf{JAPE (Ours)}}
& \textbf{\PH{44.9}} & \textbf{\PH{47.0}} & \textbf{\PH{55.1}} & \textbf{\PH{63.7}} & \textbf{\PH{54.8}} & \textbf{\PH{57.9}} & \PH{67.4} & \textbf{\PH{84.2}} & \textbf{\PH{76.6}} & \textbf{\PH{84.6}} & \textbf{\PH{59.8}} & \textbf{\PH{67.5}} \\
\bottomrule
\end{tabular}
}
\endgroup
\end{table*}

\subsection{Experimental Setup}
\label{sec:exp-setup}

\noindent\underline{\textbf{Datasets.}}
We evaluate \textbf{JAPE} on five public multivariate time-series benchmarks:
SMD~\cite{su2019omnianomaly}, WADI~\cite{ahmed2017wadi},
MSL~\cite{hundman2018spacecraft}, PSM~\cite{abdulaal2021practical},
and EXATHL-ON~\cite{jacob2021exathlon}.
All five benchmarks provide ground truth anomaly labels for the test sets, which we use directly in our experiments. SMD and WADI additionally provides event-associated
variable annotations and is therefore used for the quantitative
evaluation of variable-level explanations. Additional dataset details are provided in Appendix~\ref{app:datasets}.

\noindent\underline{\textbf{Baselines.}}
We compare \textbf{JAPE} with three categories of baselines, including unsupervised,
self-supervised, and supervised methods. The unsupervised baselines include
FCM~\cite{fcm2025} and RED-F~\cite{redf2025}. The self-supervised baseline is
A2P~\cite{park2025a2p}. For supervised comparison, we include
forecasting--detection pipelines built upon PatchTST~\cite{nie2023patchtst}
and iTransformer~\cite{liu2024itransformer}, as well as a supervised variant
of A2P (A2P-Sup) trained with real anomaly labels. Detailed descriptions of all
baselines are provided in Appendix~\ref{app:baselines}.

\noindent\underline{\textbf{Metrics.}}
We evaluate anomaly prediction using F1 and AUC-PR~\cite{davis2006relationship}.
{F1 is computed as $\mathrm{F1}=2PR/(P+R)$, where $P$ and $R$
denote point-wise precision and recall. We adopt strict point-wise matching,
where a prediction is considered correct only if it matches the anomaly label
at the same timestamp, without point adjustment~\cite{xu2018donut} or tolerance~\cite{tatbul2018precision}. AUC-PR measures
the area under the precision--recall curve and evaluates anomaly-score
discrimination across different thresholds. Both metrics are computed using
unadjusted point-wise anomaly scores.}

For variable-level explanation performance, we report HR@1, HR@3, HR@5, and MRR. HR@$k$ measures the fraction of annotated anomaly events for which at least one ground-truth variable appears among the top-$k$ predicted variables, while MRR measures the average reciprocal rank of the first ground-truth variable.

\noindent\underline{\textbf{Parameter settings.}}
We use a historical window of length $L=200$ and prediction horizons $L_{\mathrm{out}}\in\{50,100,200\}$ by default. The DSTR backbone contains $N=3$ dual-axis encoder layers with a hidden dimension of $D=128$, and the input series is divided into patches of {length $L_p=16$ with stride $S_p=8$. The alerting head is trained with focal loss~\cite{lin2017focal}, with the balancing factor set to $\alpha=0.25$ and the focusing parameter set to $\gamma=2$, respectively.} Additional implementation details and method-specific hyperparameters are provided in Appendix~\ref{app:implementation}. The code of \textbf{JAPE} is available for reproducibility\footnote{{https://anonymous.4open.science/r/JAPE-B74E/}}.


\subsection{Anomaly Prediction Performance}
\label{sec:exp-main}

Table~\ref{tab:main} reports the anomaly prediction results of \textbf{JAPE} across five datasets and three prediction lengths. Overall, \textbf{JAPE} achieves the best average performance across all settings, reaching an average F1 of 62.9 and an AUC-PR of 70.9, which outperform the strongest external baseline by 10.4 F1 points (19.7\%) and 20.7 AUC-PR points (41.3\%), respectively. The improvements are particularly pronounced on datasets with complex dependency evolution patterns. For example, on MSL, forecasting-based methods, including PatchTST, iTransformer, and A2P-Sup, achieve at most 31.8 AUC-PR, whereas \textbf{JAPE} reaches 68.1. This is because future numerical deviations alone are insufficient to capture weak anomaly precursors, whereas \textbf{JAPE} introduces additional structural anomaly evidence by jointly modeling future numerical evolution and variable dependency changes, enabling more robust anomaly identification across different prediction horizons and dataset characteristics.

In addition, the improvement of \textbf{JAPE} on AUC-PR is consistently larger than that on F1. On average, \textbf{JAPE} improves the strongest baseline by 20.7 AUC-PR points (41.3\%), compared with 10.4 F1 points (19.7\%). This gap is also evident in individual cases. For example, when $L_{out}=50$, \textbf{JAPE} achieves an AUC-PR of 62.8, substantially outperforming A2P-Sup (44.9), while their F1 scores are relatively closer. The larger gains in AUC-PR indicate that \textbf{JAPE} learns a more discriminative anomaly scoring space rather than merely optimizing performance at a specific operating point. Specifically, the jointly predicted dynamic dependency structures enhance the distinction between anomalous and normal states, allowing anomalous timestamps to receive higher and more consistent anomaly scores. Although A2P-Sup achieves slightly higher F1 under two shorter prediction lengths, \textbf{JAPE} consistently obtains substantially higher AUC-PR, e.g., 62.8 versus 44.9 when $L_{out}=50$. This demonstrates that the advantage of A2P-Sup mainly comes from its performance under a specific threshold selection, whereas \textbf{JAPE} learns a more reliable global ranking of anomaly likelihoods. Therefore, the smaller F1 gap does not imply inferior anomaly discrimination capability of \textbf{JAPE}, but rather reflects the sensitivity of threshold-based evaluation to the selected operating point.

\begin{table}[!t]
\centering
\caption{Variable-level explanation performance on SMD and WADI.}
\vspace{-3mm}
\label{tab:rca}
\small
\setlength{\tabcolsep}{2.8pt}
\renewcommand{\arraystretch}{1.08}

\begin{tabular*}{\columnwidth}{
@{\extracolsep{\fill}}
c|c@{\hspace{12pt}}|cccc
@{}}
\toprule
Dataset & Method & HR@1 & HR@3 & HR@5 & MRR \\
\midrule

\multirow{5}{*}{SMD}
& Random
& 0.177
& 0.407
& 0.557
& 0.352 \\

& Pred-Dev
& 0.165
& 0.303
& 0.416
& 0.298 \\

& GRAD
& 0.125
& 0.385
& 0.520
& 0.312 \\

& CF
& 0.235
& 0.385
& 0.486
& 0.364 \\

& \textbf{JAPE (Ours)}
& \textbf{0.333}
& \textbf{0.505}
& \textbf{0.584}
& \textbf{0.461} \\

\midrule

\multirow{5}{*}{WADI}
& Random
& 0.012
& 0.035
& 0.058
& 0.064 \\

& Pred-Dev
& 0.133
& 0.333
& 0.400
& 0.278 \\

& GRAD
& 0.000
& 0.000
& 0.067
& 0.066 \\

& CF
& \textbf{0.333}
& 0.333
& 0.400
& 0.380 \\

& \textbf{JAPE (Ours)}
& \textbf{0.333}
& \textbf{0.400}
& \textbf{0.467}
& \textbf{0.385} \\

\bottomrule
\end{tabular*}
\vspace{-3mm}
\end{table}

\subsection{Native Predictive Explanation Evaluation}
\label{sec:exp-npe}




We evaluate variable-level explanation on SMD {and WADI} datasets using their released event-associated variable annotations. We compare \textbf{JAPE} with four baselines: Random, Pred-Dev~\cite{maya2019dlstm}, GRAD~\cite{simonyan2014deep}, and CF~\cite{wachter2018counterfactual,wang2023counterfactual}. They respectively provide random ranking, prediction-error-based ranking, gradient-based saliency, and counterfactual attribution. All methods produce variable rankings for alerted events, which are evaluated against the annotated variables. 


Compared with the strongest baseline on SMD, \textbf{JAPE} improves HR@1 from 0.235 to 0.333, HR@3 from 0.407 to 0.505, HR@5 from 0.557 to 0.584, and MRR from 0.364 to 0.461, corresponding to a 26.6\% relative improvement in MRR. {Despite WADI containing only 15 events, the improvements in HR@3 and HR@5 indicate that \textbf{JAPE} more consistently ranks the attacked variables within the top candidates.} This advantage comes from the fact that \textbf{JAPE}, through its GDS module, exploits evolving dependency structures as predictive evidence. Unlike Pred-Dev, which only captures future numerical deviations, and attribution-based methods, which infer variable importance from individual model responses, \textbf{JAPE} identifies variables whose roles in system-wide dependency structures change before anomalies fully manifest. These structural changes provide complementary early signals for distinguishing influential variables from correlated but irrelevant ones, leading to more accurate variable ranking. Therefore, \textbf{JAPE} provides native predictive variable-level explanations at alert time without requiring future observations, additional post-hoc diagnosis models, or extra inference stages.

\subsection{Ablation Study}
\label{sec:exp-abl}

\begin{table}[t]
\centering
\caption{Ablation study results}
\vspace{-3mm}
\label{tab:abl}
\small
\setlength{\tabcolsep}{2.8pt}
\renewcommand{\arraystretch}{1.08}

\begin{tabular*}{\columnwidth}{
@{\extracolsep{\fill}}c|c|ccccc@{}}
\toprule
\multicolumn{2}{c|}{Method}
& SMD & WADI & MSL & PSM & EXATHLON \\
\midrule

\multirow{2}{*}{w/o DSTR}
& PatchTST
& 47.0
& 45.6
& 51.4
& \underline{78.0}
& 74.8 \\

\cmidrule(lr){2-7}

& iTransformer
& 47.5
& 11.3
& 29.1
& 76.9
& 75.2 \\

\midrule
\multicolumn{2}{c|}{w/o Dynamic dependency graph}
& \underline{51.2}
& \textbf{50.4}
& \underline{59.3}
& 77.4
& \underline{75.3} \\

\midrule

\multicolumn{2}{c|}{\textbf{JAPE (Ours)}}
& \textbf{52.5}
& \underline{49.4}
& \textbf{61.2}
& \textbf{79.0}
& \textbf{82.7} \\

\bottomrule
\end{tabular*}
\vspace{-4mm}
\end{table}



Table~\ref{tab:abl} presents the ablation study of two key components in
\textbf{JAPE} under $L_{out}=100$: the Decoupled Spatio-Temporal Representation
(DSTR) backbone and the dynamic dependency graph. 

To evaluate the effectiveness of DSTR, we replace it with two representative forecasting backbones, PatchTST and iTransformer, while keeping the alerting mechanism, training objective, and evaluation protocol unchanged. Since these variants only provide future numerical forecasts to the alerting head, this comparison directly reflects the quality of learned representations. DSTR achieves an average F1 of 62.7\%, outperforming PatchTST and iTransformer by 3.3\% and 14.7\%, respectively. Among the three backbone variants, DSTR achieves the best performance
on four of the five datasets and remains close to PatchTST on PSM. These results demonstrate that explicitly decoupling temporal evolution from inter-variable dependency modeling enables more informative representations for future anomaly prediction.

Based on DSTR, \textbf{JAPE} further incorporates the dynamic dependency graph into the alerting mechanism. This improves the average F1 from 62.7\% to 65.0\% and achieves the best performance on four out of five datasets, with the largest gain of 7.4\% on EXATHLON. On WADI, \textbf{JAPE} shows a slight decrease of 1.0\%. WADI contains 123 variables, resulting in a substantially larger candidate dependency space than the other datasets. The estimated graph may therefore retain more weak or redundant relations, slightly diluting the useful structural evidence during dual-view fusion. {Nevertheless, the overall improvement across datasets demonstrates that the dual-view alerting mechanism effectively fuses two complementary sources of information: future numerical forecasts from the temporal branch capture point-wise deviations, while evolving dependency structures from the spatial branch reveal relational shifts. By leveraging both perspectives, \textbf{JAPE} provides richer evidence beyond what numerical deviations alone can offer, leading to more robust and generalizable anomaly prediction.}

\subsection{Training Efficiency}
\label{sec:efficiency}

\begin{figure}[!t]
    \centering

    \begin{subfigure}[t]{0.49\columnwidth}
        \centering
        \includegraphics[
            width=\linewidth
        ]{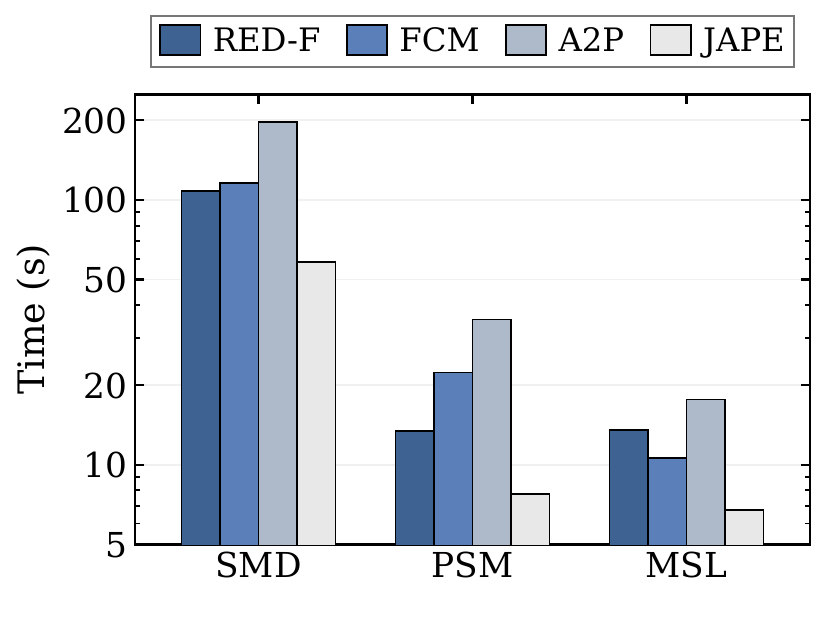}
        \caption{Training time per epoch}
        \label{fig:efficiency_epoch}
    \end{subfigure}
    \hfill
    \begin{subfigure}[t]{0.49\columnwidth}
        \centering
        \includegraphics[
            width=\linewidth
        ]{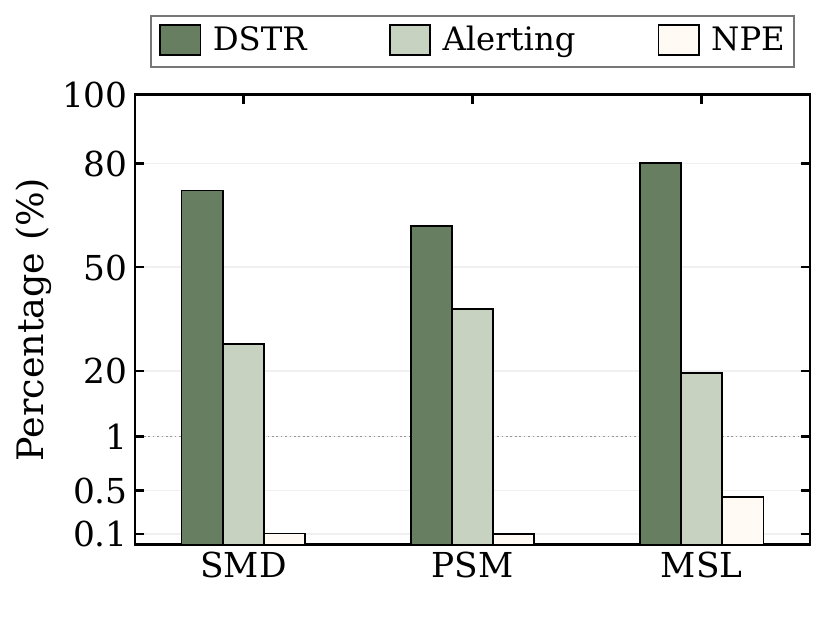}
        \caption{JAPE runtime breakdown}
        \label{fig:efficiency_components}
    \end{subfigure}
    \caption{Efficiency evaluation.}
    \label{fig:efficiency}
     \vspace{-3mm}
\end{figure}

We compare \textbf{JAPE} with A2P, FCM, and RED-F on SMD, PSM, and MSL to evaluate training efficiency. As shown in Figure~\ref{fig:efficiency}(a), \textbf{JAPE} achieves the lowest per-epoch training time on all three datasets, requiring only 58.1s, 7.77s, and 6.74s on SMD, PSM, and MSL, respectively. On SMD, \textbf{JAPE} reduces the training cost by 70.5\% compared with A2P (196.9s) and achieves nearly 2$\times$ acceleration over FCM (115.7s) and RED-F (108.3s). Similar improvements are observed on PSM and MSL, demonstrating consistent efficiency advantages across datasets. These results indicate that \textbf{JAPE} improves anomaly prediction performance without introducing additional training overhead. This efficiency is mainly attributed to the integration of dependency modeling into the forecasting backbone, where structural learning reuses the shared sequence representations and computations rather than introducing an independent modeling branch.

Figure~\ref{fig:efficiency}(b) further decomposes the runtime of \textbf{JAPE} into DSTR, Alerting, and NPE. Under the same profiling protocol, DSTR accounts for 62.0\%--80.2\% of the total runtime, while Alerting accounts for 19.4\%--37.9\%. DSTR exhibits the highest share, which is attributed to its core responsibilities: it performs the main sequence forecasting task and repeatedly constructs dynamic dependency graphs across temporal patches and encoder layers, including the future-oriented branch. These operations involve patch embedding, graph construction, and multi-head attention computations at each layer, which scale with both sequence length and encoder depth. In contrast, Alerting operates on the predicted sequence and compact structural descriptors after the backbone is frozen, requiring only lightweight standardization and neural network inference. NPE accounts for less than 0.5\% on every dataset, because it directly reuses the already generated graphs and computes deviation scores without requiring additional forward passes or auxiliary networks, confirming its negligible computational overhead.

\section{Related Work}
\label{sec:related}

\textbf{Time series anomaly detection.} Anomaly detection on multivariate time series has been extensively
studied through stochastic reconstruction
(OmniAnomaly~\cite{su2019omnianomaly}), adversarial reconstruction
(USAD~\cite{audibert2020usad}), association-discrepancy attention
(Anomaly Transformer~\cite{xu2022anomalytransformer}), and contrastive
representation learning (DCdetector~\cite{yang2023dcdetector}).
Since these methods identify anomalies from already observed data,
their alerts remain reactive. To better capture complex interactions among variables, recent studies have explored graph-based time-series modeling. GDN~\cite{deng2021gdn} and MTAD-GAT~\cite{zhao2020mtadgat} learn graph-based representations for anomaly detection, while CrossGNN~\cite{huang2023crossgnn},
CATCH~\cite{wu2025catch}, TimeFilter~\cite{hu2025timefilter}, and DyCAST~\cite{cheng2025dycast} model cross-variable or dynamic dependencies for general time-series analysis. Meanwhile, some anomaly detection methods provide variable-level interpretations through importance scores
(OmniAnomaly~\cite{su2019omnianomaly}, InterFusion~\cite{li2021interfusion}), and root-cause analysis methods further infer causal or dependency structures from observed anomaly intervals (AERCA~\cite{han2025aerca}, GCAD~\cite{liu2025gcad}, and causality-aware contrastive methods~\cite{kim2025carots}). However, these methods either detect anomalies after they occur or perform post-hoc diagnosis, and therefore cannot exploit evolving dependency changes as predictive evidence before anomaly occurrence.

\noindent \textbf{Time series anomaly prediction.} Anomaly prediction instead aims to identify abnormal time steps over an
explicit future horizon. Early work detects precursors indicating an
imminent anomaly (PoA~\cite{jhin2023pad}), while subsequent work
formalizes point-wise anomaly prediction over future sequences
\cite{you2024anomaly}. Later methods introduce anomaly-related information at different stages of the forecasting-based prediction process. TranAP~\cite{tfattn2025} forecasts future series and evaluates them through reconstruction errors learned from normal data. FCM~\cite{fcm2025} uses forecasted future context to amplify weak precursors, whereas RED-F~\cite{redf2025} contrasts forecasts generated from the original sequence and a reconstructed normal-pattern sequence. A2P~\cite{park2025a2p} introduces anomaly-aware forecasting and synthetic anomaly prompting, MultiRC~\cite{hu2024multirc} strengthens anomaly-related representations through multi-scale reconstructive contrast, and F2A~\cite{f2a2025} adapts time-series foundation models through joint forecast--anomaly optimization and relevant-horizon retrieval.

Despite their different mechanisms, existing anomaly prediction methods
primarily derive future alerts from numerical signal patterns or their
latent representations. They do not explicitly preserve evolving
directed inter-variable dependencies as predictive evidence. \textbf{JAPE}
addresses this limitation by jointly modeling future numerical evolution
and dynamic dependency structures, combining both views for point-wise
anomaly prediction and reusing the predicted dependency graphs for
native variable-level explanation.
\section{Conclusion}
\label{sec:conclusion}

In this paper, we present \textbf{JAPE}, a joint anomaly prediction and explanation framework that lifts anomaly prediction from numerical-deviation modeling to dependency-structure modeling. \textbf{JAPE} constructs a dynamic directed dependency-graph sequence spanning the observed history and the predicted future through the Decoupled Spatio-Temporal Representation backbone (DSTR), drives point-wise alerting with two complementary views---numerical and structural---and reuses the same graph through Native Predictive Explanation (NPE) to deliver variable-level explanations within the same forward pass via the Graph Deviation Score. {Experiments on five real-world benchmarks demonstrate that \textbf{JAPE} achieves competitive prediction performance under a strict point-wise evaluation protocol, while producing explanations at negligible additional cost. In the future, we plan to scale dependency-structure modeling to larger systems with hundreds of variables, explore adaptive graph sparsification strategies to reduce computational overhead, and investigate the potential of structural signals in online continual-learning scenarios where anomaly patterns evolve over time.}



\nocite{li2020compression, li2021trace, li2022evolutionary, hu2023spatio, yao2024camel, yao2025moon}
\balance
\bibliographystyle{ACM-Reference-Format}
\bibliography{sample}

\clearpage
\appendix
\clearpage
\appendix

\twocolumn[
\begin{minipage}{\textwidth}
\centering

\captionof{table}{Dataset statistics used in the experiments.}
\label{tab:dataset-statistics}

\setlength{\tabcolsep}{5pt}
\begin{tabular}{lrrrrrrr}
\toprule
Dataset &
Series &
Variables &
Train Points &
Test Points &
Anomaly Points &
Anomaly Ratio (\%) &
Anomaly Events \\
\midrule
SMD      & 28 & 38  & 708405  & 708420 & 29444 & 4.16  & 327 \\
WADI     & 1  & 123 & 1209601 & 172801 & 9985  & 5.78  & 15  \\
MSL      & 1  & 55  & 58317   & 73729  & 7726  & 10.53 & 36  \\
PSM      & 1  & 25  & 132481  & 87841  & 24381 & 27.76 & 72  \\
EXATHLON & 8  & 19  & 1232389 & 660442 & 84061 & 12.73 & 86  \\
\bottomrule
\end{tabular}

\vspace{1em}
\end{minipage}
]

\section{Dataset Statistics}
\label{app:datasets}

We evaluate JAPE on five widely used multivariate time-series benchmarks: SMD, WADI, MSL, PSM, and EXATHLON. These datasets represent diverse real-world scenarios, including server monitoring, spacecraft telemetry, industrial control systems, and large-scale computing platforms, with varying numbers of variables, temporal scales, and anomaly characteristics. Table~ \ref{tab:dataset-statistics} summarizes their dataset statistics and anomaly distributions. For datasets containing multiple independent time series, the reported training and testing lengths are aggregated across all series, while anomaly events are counted separately within each individual series.

We follow the finalized data-loading pipeline for each benchmark and use the
provided train--test partitions. Data-dependent preprocessing statistics are
estimated from the training split only. Historical windows of length $L$ are
used to predict point-wise anomaly probabilities over the subsequent horizon
$H$. Validation data are used for model selection, early stopping, and
threshold selection, while the test split is reserved for final evaluation.


\section{Baseline Details}
\label{app:baselines}
We compare \textbf{JAPE} against three classes of baselines.
\begin{itemize}[topsep=0pt,itemsep=0pt,parsep=0pt,partopsep=0pt,leftmargin=*]

\item {Unsupervised baselines.}
FCM~\cite{fcm2025} leverages the forecasted future context to amplify
weak anomaly precursors in the observation window.
RED-F~\cite{redf2025} further contrasts the forecasts derived from the
original sequence and a reconstructed normal-pattern baseline to
highlight subtle anomaly precursors.

\item {Self-supervised baseline.}
A2P~\cite{park2025a2p} injects multi-form pseudo-anomalies through a
learnable prompt pool to learn anomaly-related patterns in a
self-supervised manner. We retain the original A2P in the training-efficiency
comparison.

\item {Supervised baselines.} We construct forecasting--detection pipelines
by augmenting PatchTST~\cite{nie2023patchtst} and
iTransformer~\cite{liu2024itransformer} with
AT~\cite{xu2022anomalytransformer} or CAD~\cite{cad_ref} as detection
heads. For a directly comparable supervised anomaly-prediction setting,
we adapt A2P as A2P-Sup by training it with real anomaly labels rather
than injected pseudo-anomalies. All supervised methods share identical
labels and evaluation protocols.
\end{itemize}

\section{Implementation Details}
\label{app:implementation}

The forecasting backbone DSTR is trained with an input length of
$L=200$ and evaluated under three prediction horizons,
$H=L_{\mathrm{out}}\in\{50,100,200\}$. The input sequence is divided
into patches of length $L_p=16$ with stride $S_p=8$. DSTR consists of
three encoder layers with a hidden dimension of 128, eight attention
heads, an FFN dimension of 256, and a dropout rate of 0.1. For each
target variable, we retain the top $K_g=5$ predecessor variables, with
a maximum lag range of $K_{\max}=3$ for all datasets except WADI,
where $K_{\max}=5$. The residual graph coefficient $\lambda$ is
jointly optimized with the forecasting backbone. RevIN with affine
transformation is applied before forecasting. The backbone is
optimized using Adam with a batch size of 128, a learning rate of
$10^{-4}$, and a maximum of 20 epochs with early stopping
(patience 5).

For structural feature construction, each historical or future graph
segment is divided into an earlier portion and a later portion, where
the later portion contains the final $\tau=0.3$ fraction of graph
slices. Segment-level changes are computed from the earlier portion to
the later portion.

The alerting head contains three layers with a hidden dimension of 64
and four attention heads. It is optimized using focal loss with
$\alpha=0.25$ and $\gamma=2$, using the same batch size and learning
rate as the backbone, together with a weight decay of $10^{-4}$.
Training is performed for at most 20 epochs with early stopping
(patience 5). 

For the proposed NPE module, we set $\beta=0.7$,
$K_{\mathrm{path}}=2$, $\omega=1.0$,
$\alpha_{\mathrm{gds}}=0.8$, and $R=1$. The normal reference graph
and the variable-wise statistics used to standardize the direct and
multi-hop deviation scores are estimated from normal reference
windows. Variable-level explanations use only the predicted graph
slices associated with the alerted segment. Complete dataset-specific
configurations and reproduction scripts are provided in the
anonymized repository.

\begin{algorithm}[t]
\caption{Two-stage training and inference of \textbf{JAPE}.}
\label{alg}
\LinesNumbered

\KwIn{
Forecast-training windows
$\mathcal{D}=\{(\mathbf{X},\mathbf{X}^{+})\}$;
labeled windows
$\mathcal{D}_{\ell}=\{(\mathbf{X},\mathbf{y})\}$;
normal reference windows $\mathcal{D}_{n}$;
inference window $\mathbf{X}_{*}$;
alert threshold $\eta$
}

\KwOut{
Point-wise anomaly probabilities $\mathbf{p}$;
variable-level explanation scores $\mathbf{s}$
}

\tcp{Stage 1: forecast-oriented backbone training}
\ForEach{$(\mathbf{X},\mathbf{X}^{+})\in\mathcal{D}$}{

$(\mathbf{Z},
\mathbf{A}_{\mathrm{hist}},
\mathbf{A}_{\mathrm{hist}}^{\mathrm{raw}})
\leftarrow
\textsc{SpatialTemporalAxes}(\mathbf{X})$\;

$\widehat{\mathbf{X}}
\leftarrow
\textsc{ForecastHead}(\mathbf{Z})$\;

$(\widehat{\mathbf{X}}_{\mathrm{fut}},
\widehat{\mathbf{A}}_{\mathrm{fut}},
\widehat{\mathbf{A}}_{\mathrm{fut}}^{\mathrm{raw}})
\leftarrow
\textsc{FutureOrientedBranch}
(\operatorname{sg}(\mathbf{Z}))$\;

$\mathbf{A}
\leftarrow
[\mathbf{A}_{\mathrm{hist}};
 \widehat{\mathbf{A}}_{\mathrm{fut}}],
\quad
\mathbf{A}^{\mathrm{raw}}
\leftarrow
[\mathbf{A}_{\mathrm{hist}}^{\mathrm{raw}};
 \widehat{\mathbf{A}}_{\mathrm{fut}}^{\mathrm{raw}}]$\;

Update the historical backbone and forecasting head using
$\mathcal{L}_{\mathrm{MSE}}
(\widehat{\mathbf{X}},\mathbf{X}^{+})$,
and the future-oriented branch using
$\mathcal{L}_{\mathrm{MSE}}^{\mathrm{fut}}
(\widehat{\mathbf{X}}_{\mathrm{fut}},\mathbf{X}^{+})$\;
}

Freeze the forecasting and graph-construction modules\;

\tcp{Stage 2: supervised alert-head training}
\ForEach{$(\mathbf{X},\mathbf{y})\in\mathcal{D}_{\ell}$}{

$(\widehat{\mathbf{X}},
\mathbf{A},
\mathbf{A}^{\mathrm{raw}})
\leftarrow
\textsc{FrozenBackbone}(\mathbf{X})$\;

$\mathbf{p}
\leftarrow
\textsc{AlertHead}
(\widehat{\mathbf{X}},
\mathbf{A},
\mathbf{A}^{\mathrm{raw}})$\;

Update the alerting head using
$\mathcal{L}_{\mathrm{focal}}
(\mathbf{p},\mathbf{y})$\;
}

Estimate
$\mathbf{A}^{\mathrm{normal}}$
and the variable-wise normalization statistics
$\mathcal{S}_{n}$
from $\mathcal{D}_{n}$ using the frozen backbone\;

\tcp{Inference}
$(\widehat{\mathbf{X}}_{*},
\mathbf{A}_{*},
\mathbf{A}^{\mathrm{raw}}_{*})
\leftarrow
\textsc{FrozenBackbone}(\mathbf{X}_{*})$\;

$\mathbf{p}
\leftarrow
\textsc{AlertHead}
(\widehat{\mathbf{X}}_{*},
\mathbf{A}_{*},
\mathbf{A}^{\mathrm{raw}}_{*})$\;

$\mathbf{s}\leftarrow\varnothing$\;

\If{$\max_{h}p_{h}\geq\eta$}{

$\mathbf{A}^{\mathrm{anom}}
\leftarrow
\textsc{AggregateAlertedGraphs}
(\mathbf{A}_{*},\mathbf{p},\eta)$\;

$\mathbf{s}
\leftarrow
\textsc{GDS}
(\mathbf{A}^{\mathrm{anom}},
\mathbf{A}^{\mathrm{normal}},
\mathcal{S}_{n})$\;
}

\Return{$\mathbf{p},\mathbf{s}$}\;

\end{algorithm}

\section{Training and Inference Procedure}
\label{app:training}

Algorithm~\ref{alg} summarizes the two-stage training and inference procedure of \textbf{JAPE}. Given forecast-training windows $\mathcal{D}$, labeled windows $\mathcal{D}{\ell}$, normal reference windows $\mathcal{D}{n}$, an inference window $\mathbf{X}_{*}$, and an alert threshold $\eta$, \textbf{JAPE} outputs point-wise anomaly probabilities $\mathbf{p}$ and variable-level explanation scores $\mathbf{s}$.

The training procedure consists of two stages. In Stage~1, \textbf{JAPE} trains the forecasting-oriented backbone to learn numerical evolution and dynamic dependency structures. For each forecast-training pair $(\mathbf{X},\mathbf{X}^{+})$, DSTR extracts historical representations and dependency graphs, while the forecasting head and future-oriented branch generate numerical forecasts and future dependency graphs (lines~1--5). The historical and future graphs are combined to characterize dependency evolution across the observation and prediction horizons. The backbone and future-oriented branch are optimized with forecasting objectives, and the forecasting and graph-construction modules are then frozen (lines~6--7). In Stage~2, the frozen backbone produces forecasts and graph sequences for labeled windows, and the alerting head jointly integrates numerical and structural information to estimate anomaly probabilities (lines~8--10). The alerting head is optimized with focal loss, while normal graph statistics and variable-wise normalization factors are extracted from $\mathcal{D}_{n}$ for subsequent explanation (lines~11--12).

During inference, \textbf{JAPE} first feeds the input window $\mathbf{X}{*}$ into the frozen backbone to obtain the future forecasts and dependency graph sequences, which are then passed to the alerting head to compute point-wise anomaly probabilities (lines~13--14). If the maximum anomaly probability is below the threshold $\eta$, the explanation score $\mathbf{s}$ remains empty. Otherwise, \textbf{JAPE} aggregates the graph slices corresponding to the detected anomaly horizon into $\mathbf{A}^{\mathrm{anom}}$ and applies GDS with the normal reference graph $\mathbf{A}^{\mathrm{normal}}$ and normalization statistics $\mathcal{S}{n}$ to compute variable-level explanation scores (lines~15--18). Finally, JAPE returns the anomaly probabilities $\mathbf{p}$ and explanation scores $\mathbf{s}$ (line~19).

\section{Sensitivity Analysis}
\label{app:sens}
Tables~\ref{tab:sens_detection} and~\ref{tab:sens_explanation} report the sensitivity results of DSTR and NPE, respectively.

For DSTR, $K_{\max}$ shows a clear impact on performance. Increasing $K_{\max}$ from 1 to 3 consistently improves the results on SMD and MSL, with F1 increasing from 50.90 to 52.50 and from 55.50 to 61.20, respectively. This indicates that incorporating multiple lag steps helps capture delayed dependency evolution that cannot be identified from immediate interactions alone. However, further increasing $K_{\max}$ to 5 causes a significant performance drop, especially on MSL, where F1 decreases from 61.20 to 35.50. This is because overly long lag 
ranges may introduce weak or spurious dependencies, making the learned dependency graph less discriminative. The optimal $K_g$ varies across datasets, reflecting different dependency densities among variables. For example, a larger neighborhood benefits SMD, while smaller neighborhoods perform better on MSL and PSM. This suggests that excessive aggregation of source variables may introduce redundant information and dilute informative dependency signals. The default setting ($K_g=5$) provides a balanced trade-off between dependency coverage and noise.

\begin{table}[th]
\centering
\small
\setlength{\tabcolsep}{3pt}

\caption{Sensitivity analysis of DSTR.}
\label{tab:sens_detection}
\vspace{-2mm}

\begin{tabularx}{\columnwidth}{
@{}
>{\centering\arraybackslash}m{0.16\columnwidth}|
>{\centering\arraybackslash}m{0.13\columnwidth}|
*{3}{>{\centering\arraybackslash}X}
@{}
}
\toprule
Param. & Value & SMD & MSL & PSM \\
\midrule

\multirow{4}{*}{$K_{\max}$}
& $1$   & $50.90$          & $55.50$          & $\textbf{82.50}$ \\
& $2$   & $51.60$          & $58.40$          & $79.80$ \\
& $3^*$ & $\textbf{52.50}$ & $\textbf{61.20}$ & $79.00$ \\
& $5$   & $46.50$          & $35.50$          & $78.40$ \\
\midrule
\multirow{5}{*}{$K_g$}
& $1$   & $47.25$          & $\textbf{65.34}$ & $\textbf{79.21}$ \\
& $3$   & $47.35$          & $51.48$          & $76.51$ \\
& $5^*$ & $52.50$          & $61.20$          & $79.00$ \\
& $10$  & $\textbf{54.01}$ & $47.39$          & $74.46$ \\
& $15$  & $53.89$          & $46.02$          & $77.44$ \\
\bottomrule
\end{tabularx}

\vspace{2mm}

\caption{Sensitivity analysis of NPE.}
\label{tab:sens_explanation}
\vspace{-2mm}

\begin{tabularx}{\columnwidth}{
@{}
>{\centering\arraybackslash}m{0.16\columnwidth}|
>{\centering\arraybackslash}m{0.13\columnwidth}|
*{4}{>{\centering\arraybackslash}X}
@{}
}
\toprule
Param. & Value & HR@1 & HR@3 & HR@5 & MRR \\
\midrule

\multirow{4}{*}{$K_{\mathrm{path}}$}
& $1$   & $32.11$          & $\textbf{51.38}$ & $58.10$          & $45.28$ \\
& $2^*$ & $\textbf{33.03}$ & $51.07$          & $\textbf{63.00}$ & $\textbf{46.50}$ \\
& $3$   & $\textbf{33.03}$ & $50.76$          & $61.77$          & $46.44$ \\
& $5$   & $32.72$          & $51.07$          & $61.77$          & $46.34$ \\
\midrule

\multirow{7}{*}{$\omega$}
& $0.00$   & $\textbf{33.03}$ & $51.07$          & $63.00$          & $47.02$ \\
& $0.25$   & $\textbf{33.03}$ & $51.07$          & $59.67$          & $46.51$ \\
& $0.50$   & $\textbf{33.03}$ & $51.07$          & $59.67$          & $46.58$ \\
& $0.75$   & $\textbf{33.03}$ & $51.07$          & $61.34$          & $46.38$ \\
& $1.00^*$ & $\textbf{33.03}$ & $51.07$          & $63.00$          & $46.50$ \\
& $1.50$   & $\textbf{33.03}$ & $\textbf{52.74}$ & $\textbf{64.67}$ & $\textbf{47.13}$ \\
& $2.00$   & $\textbf{33.03}$ & $51.07$          & $\textbf{64.67}$ & $46.98$ \\
\bottomrule
\end{tabularx}

\end{table}

For NPE, the explanation performance remains relatively stable under different parameter settings. Increasing $K_{\mathrm{path}}$ from 1 to 2 improves HR@5 and MRR (from 58.10 to 63.00 and from 45.28 to 46.50), while further increasing it brings negligible gains. This suggests that two-hop dependency paths are sufficient to capture the major propagation patterns contributing to anomalies, whereas deeper paths may introduce indirect and less reliable relationships. The weighting factor $\omega$ has limited influence on the results, with most settings achieving comparable performance. Nevertheless, $\omega=1.5$ achieves the highest HR@3, HR@5, and MRR, indicating that moderately emphasizing structural dependency information can better prioritize anomaly-related variables. Overall, these results demonstrate that NPE effectively exploits local dependency structures while maintaining robustness to parameter variations.


\end{document}